\documentclass{article}

\usepackage[final]{neurips_2022}

\usepackage[utf8]{inputenc} 
\usepackage[T1]{fontenc}    
\usepackage{hyperref}       
\usepackage{url}            
\usepackage{booktabs}       
\usepackage{amsfonts}       
\usepackage{nicefrac}       
\usepackage{microtype}      
\usepackage{xcolor}         
\usepackage{epsfig}

\newcommand{\be}{\begin{equation}}
\newcommand{\ee}{\end{equation}}

\newcommand{\ba}{\begin{array}}
\newcommand{\ea}{\end{array}}
\newcommand{\bea}{\begin{eqnarray}}
\newcommand{\eea}{\end{eqnarray}}

\newcommand{\mv}{}

\usepackage{mathtools}
\newcommand{\bs}{\boldsymbol}

\title{Multimodal Transformer for Parallel Concatenated Variational Autoencoders}

\author{%
  Stephen D. Liang\\
  Department of Computer Science\\
  University of Southern California\\
  Los Angeles, CA 90089, USA \\
  \texttt{sdliang@usc.edu} \\
   \And
   Jerry M. Mendel \\
   Department of Electrical and Computer Engineering\\
   University of Southern California\\
  Los Angeles, CA 90089, USA \\
  \texttt{mendel@sipi.usc.edu} \\
}

\begin{document}

\maketitle

\begin{abstract}
In this paper, we propose a multimodal transformer using parallel concatenated architecture. Instead of using patches, we use column stripes for images in R, G, B channels as the transformer input. The column stripes keep the spatial relations of original image.
We incorporate the multimodal transformer with variational autoencoder for synthetic cross-modal data generation. The multimodal transformer is designed using multiple compression matrices, and it
serves as encoders for Parallel Concatenated Variational AutoEncoders (PC-VAE). 
The PC-VAE consists of multiple encoders, one latent space, and two decoders.
The encoders are based on random Gaussian matrices and don't need any training. 
We propose a new loss function based on the interaction information from partial information decomposition.
The interaction information evaluates the input cross-modal information and 
decoder output. The PC-VAE are trained 
via minimizing the loss function.
Experiments are performed to validate the proposed multimodal transformer for PC-VAE.    
\end{abstract}

\section{Introduction}

Multimodal machine learning
studies models to process, generate, or integrate information from multiple modalities~\cite{Balt18}.
The availability of benchmark multimodal datasets is important for machine learning research. However, some datasets are very limited because of time (e.g., COVID-19 data in early 2020), resources (e.g., art masterpieces), ethics (e.g., colon animals) or the cost of experiments (e.g., space shuttle launching), etc.
Synthetic data generated from 
Variational AutoEncoder (VAE) and Generative Adversarial Network (GAN) 
are often used to solve the data scarcity problem.
In this paper, we focus on multimodal transformer-based VAE.

Multimodal transformer has become a hot research area. 
It is often related with natural language or text processing.
In \cite{Tsai19}, multimodal transformer was applied to 
human language with a mixture of natural language, facial gestures, and acoustic behaviors, and addressed issues such as long-range dependencies across modalities and inherent data non-alignment. 
In \cite{Hu21}, 
a Unified Transformer model was proposed to learn the tasks across different multimodal domains such as natural language understanding and multimodal reasoning. Each input modality was encoded by an encoder and and encoded outcomes were used for the predictions via a shared decoder.
In \cite{huang2021unifying},
the joint learning of image-to-text and text-to-image generations was studied 
based on a single multimodal model to target the bi-directional tasks, and a transformer was adopted with a unified architecture for the
performance and task-agnostic design. 
In \cite{zhang2022transformer},
the multimodal features such as speech prosody, verbal words, and facial expression, were extracted from the video clips for multimodal transformer.
In \cite{liu2020multi},
a Multi-Level Multi-Modal Transformer was proposed to process and integrate 
multiple textual instructions and multiple images. 
In \cite{hu2020iterative}, 
a multimodal transformer architecture was proposed along with a rich representation for text in images.
This approach targets the TextVQA task.
In \cite{kant2020spatially},
 reasoning about text in images to answer a question was studied using multimodal transformer.
In \cite{yu2020improving},
Multimodal Named Entity Recognition was proposed for social media posts.
a multimodal interaction module was introduced to get both image-related word representations and word-related visual representations.

In multimodal transformers for general studies,
multimodal self-attention in transformer was proposed to 
make unequal treatment to different modalities and avoid encoding useless information from less important modalities~\cite{Yao20}.
In \cite{Yu19},
multimodal transformer was proposed for image captioning,
which simultaneously captures 
interactions within modality and cross-modality
with multimodal reasoning and output accurate captions.
In \cite{Zadeh19},
the Factorized Multimodal Transformer was presented for multimodal sequential learning, which could increase the number of self-attentions to better model the multimodal phenomena. This model can  extract the information of long-range multimodal dynamics
asynchronously.
In \cite{chen2021history}, 
 a History Aware Multimodal Transformer was introduced to help with
 multimodal decision making
 via
 incorporating a long-horizon history. 
This model used
a hierarchical vision transformer
to
 encode all the previous observations.
In \cite{hendricks2021decoupling},
three important factors were investigated for multimodal transformer that can impact the representations of pretraining data, loss function, and the attention mechanism. 
In \cite{zhu2020enhance},
Zhu studied
multimodal transformer enhancement using external label and in-domain pretrain.
In \cite{wang2020transmodality},
a new fusion method, TransModality, was introduced to multimodal sentiment analysis.

In the theoretical and behavior studies of multimodal transformer,
 the behavior of Transformers with modal-incomplete data
 was 
 studied in \cite{ma2022multimodal}, and it was
 observed that
 transformer models are sensitive to missing modalities and different modal fusion strategies could affect the robustness.
In \cite{xie2022mnsrnet},
an effective multimodal-driven deep neural network was proposed to 
perform 3D surface super-resolution in the regular 2D domain. This model
jointly consider the normal modalities, depth, and texture.
In \cite{akbari2021vatt},
 multimodal representations learning from unlabeled data 
 was introduced using transformer architectures without using convolution. 
 In \cite{cheng2021multimodal}, 
multimodal Sparse Phased Transformer was studied to reduce the complexity of self-attention and memory footprint. 
Parameter sharing in each layer and co-attention factorization that share parameters between cross attention blocks was introduced to minimize impact on task performance.
To make effective interaction among different modalities, a universal multimodal transformer was introduced to learn joint representations among different modalities \cite{li2021bridging}.

The contributions of this paper include the following.
\begin{enumerate}
    \item 
    We propose a multimodal transformer 
     for synthetic cross-modal data generation. 
The multimodal transformer is designed using multiple compression matrices.

\item
We propose a new machine learning model, parallel concatenated variational autoencoders (PC-VAE). 
The multimodal transformer serves as encoders for PC-VAE.
The PC-VAE consists of multiple encoders, one latent space, and two decoders (visual and audio).

\item
We propose a new vision transformer. Instead of using patches, we use stripes for images in R, G, B channels.

\item
We propose a new loss function based on the interaction information from partial information decomposition.
The interaction information evaluates the input cross-modal information and 
decoder output. 
The PC-VAE are trained 
via minimizing the interaction information.

\end{enumerate}

\section{Related Work}

The vision transformer was proposed in \cite{Doso21}, which divides an image into multiple patches, and each patch can be processed individually. 
Recently, ViT~\cite{Doso21} achieved very good results 
with less computational cost.
In ViT, an image was split into fixed-size patches, linearly embedded,
and then position embeddings were added. The resulting vector was fed to a standard Transformer
encoder. To perform classification using ViT, an extra learnable
“classification token” was added to the sequence.

In \cite{He21}, masked autoencoder was proposed using vision transformer to recover the original images even if some patches are masked.
It develops an asymmetric encoder-decoder architecture.
Its encoder is a Vision Transformer (ViT)~\cite{Doso21}, and was applied only
on visible, unmasked patches.
This transformer has been very successful for 
natural language processing.
In \cite{Li21},  a conditional VAE
was proposed to
generate diverse gestures from speech audio.
The conditional VAE
was used to model
audio-to-motion mapping via decomposing the cross-modal
latent code into motion-specific code and shared code. 
Visual and audio contain both common and complementary
information. In \cite{Hao18}, a
cross-modal cycle generative adversarial network was proposed 
to handle cross-modal visual-audio mutual generation.

In the related works on multimodal transformers for visual and audio information. In \cite{zhao2022hierarchical}, 
 audio and visual information was integrated using multimodal fusion mechanism based on a hierarchical transformer,
 which could capture the dependencies among frame and shots.
 Multimodal transformer was used to fuse audio-visual modalities on the model level in \cite{Huang20}. After encoding audio and visual modalities,
the multi-head attention was applied to produce multimodal emotional intermediate representations from semantic feature space. 
In \cite{parthasarathy2021detecting},
deep-learning algorithms for audio-visual detection of user’s expression
was studied, and a transformer architecture with encoder layers was proposed to integrate audio-visual features for expressions tracking.
In \cite{Dzabraev21}, a multidomain multimodal transformer was proposed for video retrieval, which provided
multidomain generalisation via combining different video caption datasets.
In \cite{lee2020parameter}, a parameter sharing scheme based on low-rank approximation
was proposed to reduce
the number of parameters associated with the audio-visual video representation learning in
multimodal transformers. 
The Transformer was decomposed into modality-specific and modality-shared parts.
In \cite{rahman2020integrating},
multimodal adaptation gate was introduced to accept multimodal nonverbal data during fine-tuning, and the visual and acoustic modalities were studied in this model.
In \cite{le2019multimodal},
multimodal transformer networks was proposed for end-to-end video-grounded dialogue systems, and a query-aware attention through an auto-encoder 
was studied to obtain query-related features from non-text information.
In \cite{xie2021robust},
a multimodal emotion recognition algorithm during a conversation
was studied, and separate models for visual and audio were structured and trained.

In the current VAE,
Evidence Lower Bound (ELBO) function
is often used as the cost function for VAEs, and
ELBO includes Kullback–Leibler 
(KL) divergence and
reconstruction error~\cite{King14}.
The purpose of VAE is to generalize the input and provide generative modeling for the input.
In \cite{Luca19},
ELBO was used to train the linear VAEs, and 
comparing to the relative to log marginal likelihood method,
this approach
does not
introduce any additional spurious local maxima. 
In this paper, we propose a new cost function for synthetic data evaluation and performance metric using interaction information from Partial Information Decomposition (PID).

The loss function has been used for VAE training.
Evidence lower bound objective (ELBO) loss function has been widely to train the VAE parameters~\cite{King14}~\cite{Lucas19}.
In \cite{Lian21}, transfer entropy was used as a loss function for VAE training.
All the above loss functions are based on a VAE system with same modality and single input and single output system. In our multimodal transformer PC-VAE, the input has multimodality, so we propose a new loss function based on Partial Information Decomposition.

\section{Multimodal Transformer for Parallel Concatenated VAEs}

\subsection{Transformer and Architecture}

\begin{figure}[ht]
    \begin{center}
     \begin{tabular}{cc}
\epsfig{file=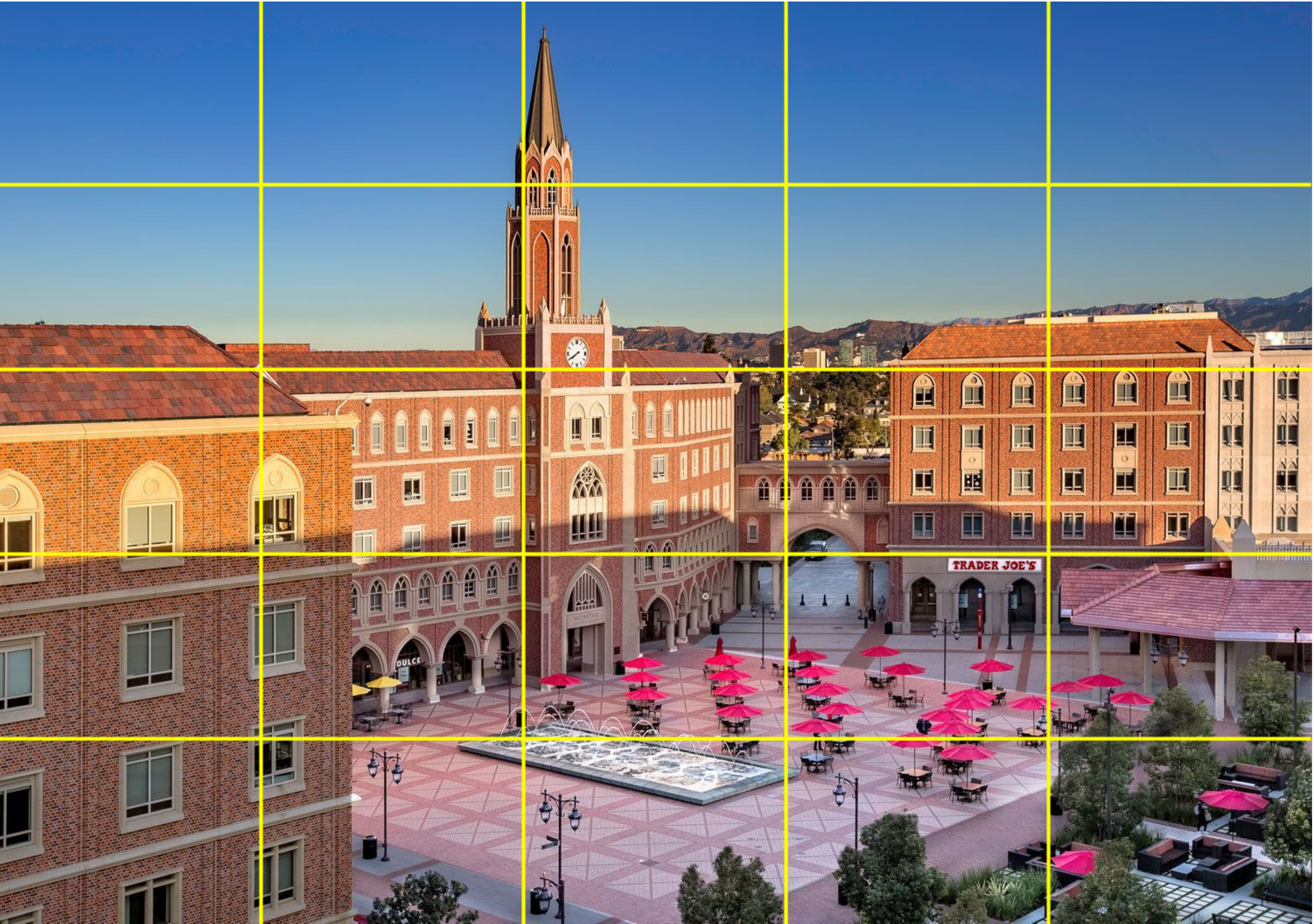,width=2.2in}  &
\epsfig{file=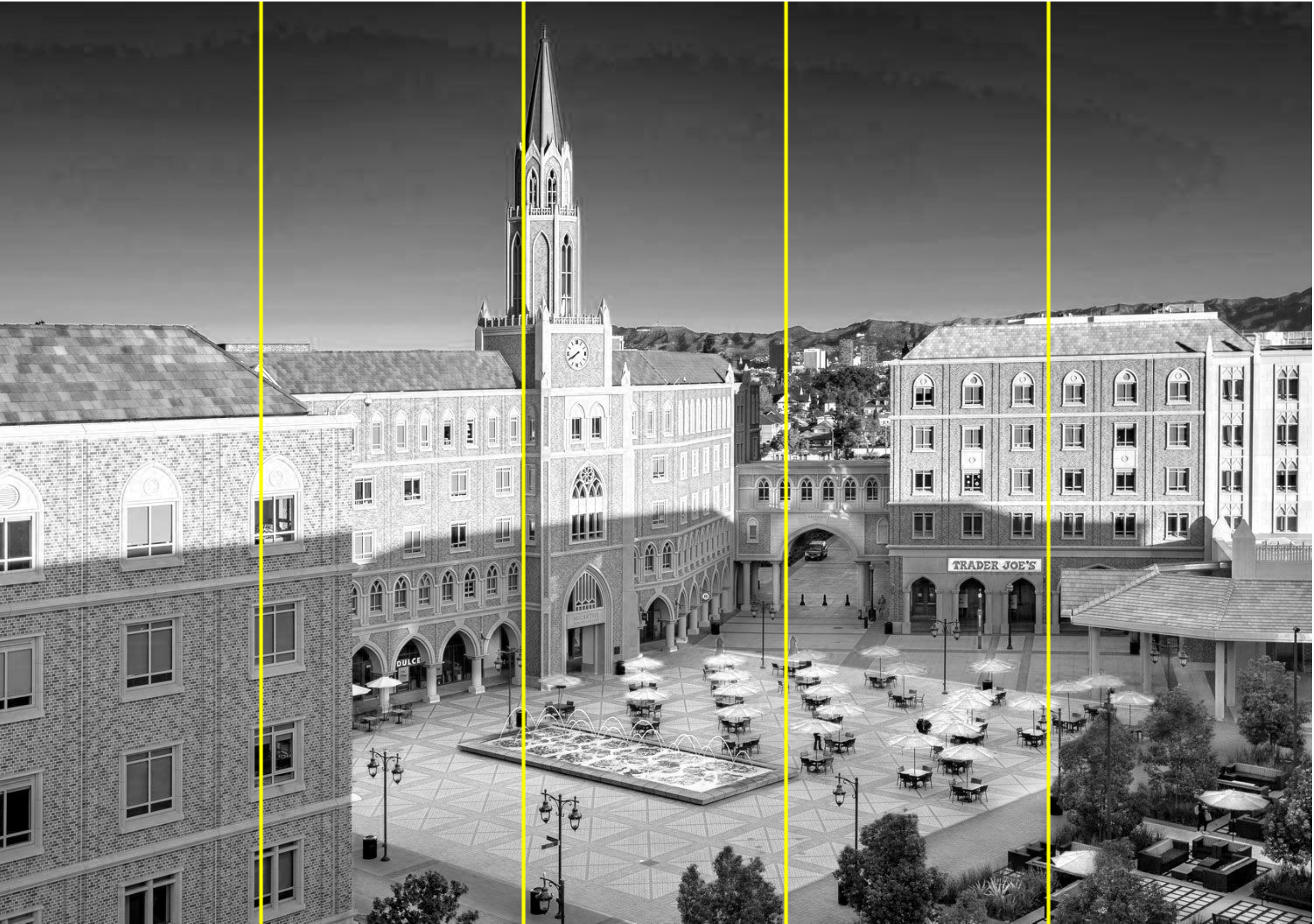,width=2.2in}  \\
(a)&(b)\\
\epsfig{file=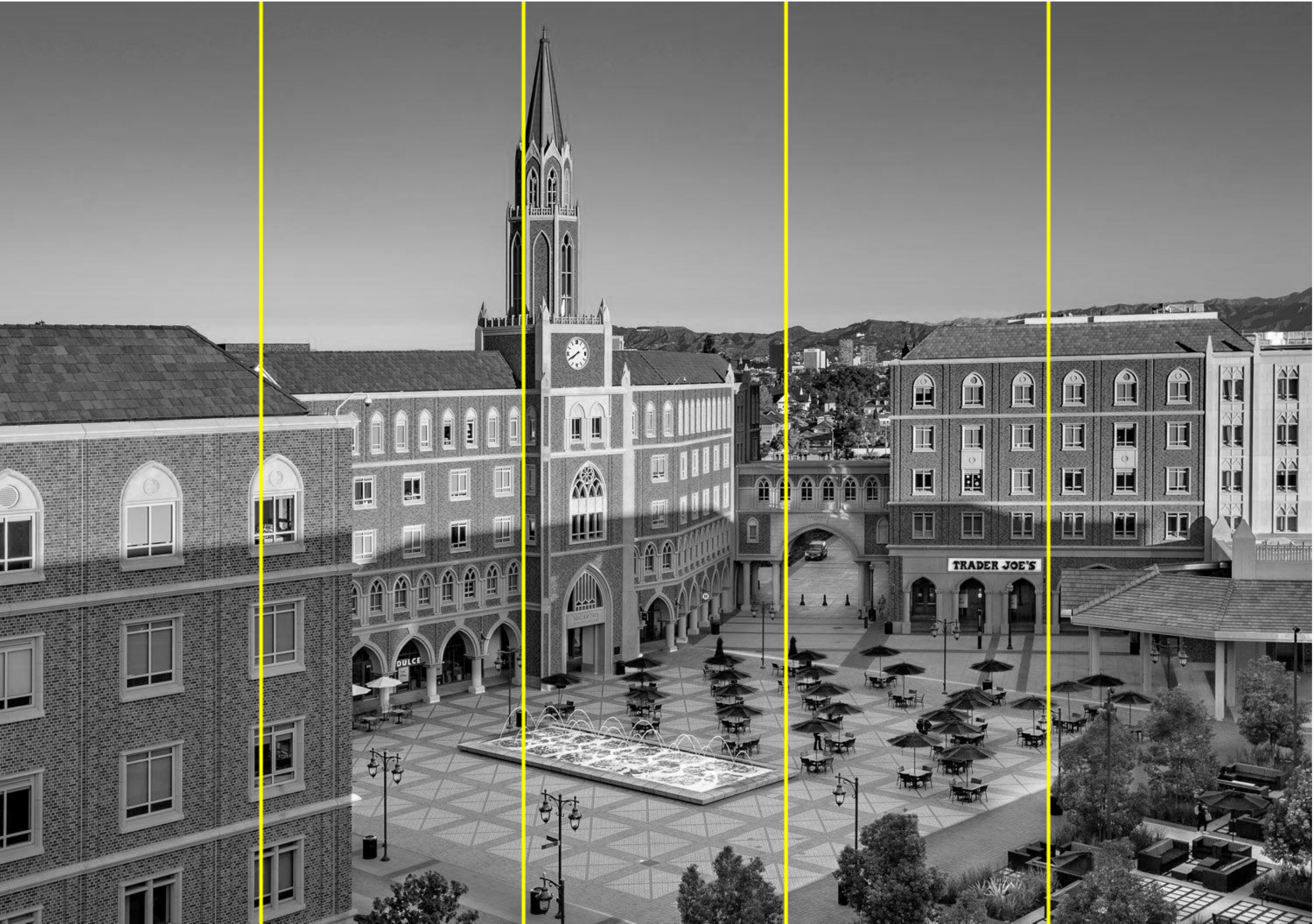,width=2.2in}  &
\epsfig{file=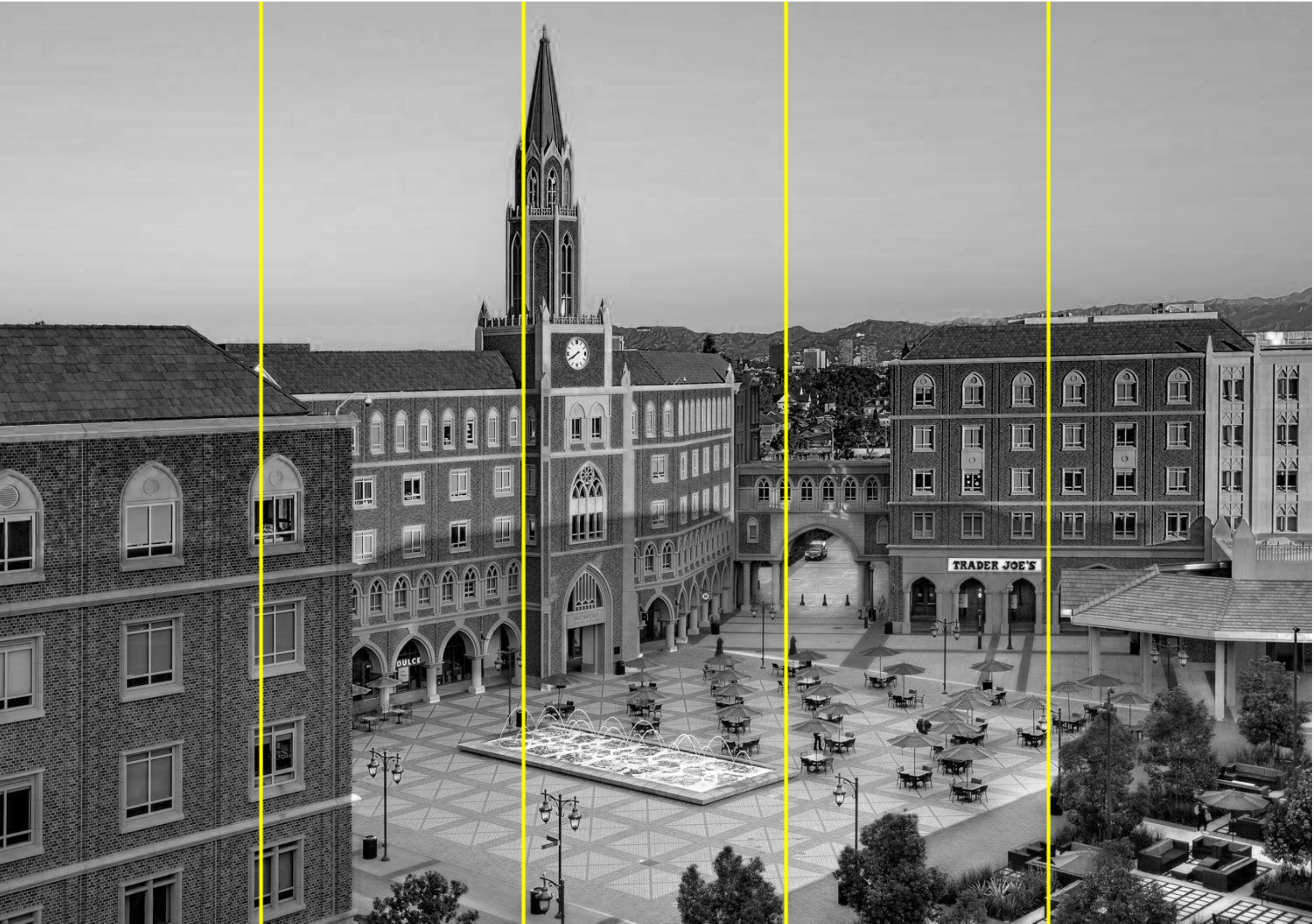,width=2.2in}  \\
(c) & (d)
 \end{tabular}
    \end{center}
    \caption{Different transformer schemes.
    (a) Existing approach. (b) Transformer for R channel. (c) Transformer for G channel. (b) Transformer for B channel.
     }
    \label{fig:0}
\end{figure}

In the existing vision transformer~\cite{Doso21}\cite{He21}, uniformly partitioned patches are used for an image, as illustrated in Fig.
\ref{fig:0}(a).
In this paper, we propose a new vision transformer using stripes
from each RGB channel image,
as illustrated in Fig. \ref{fig:0}(b)(c)(d).
Then linear transformation is applied to each stripe in the RGB channel images respectively.

\begin{figure}[ht]
    \begin{center}
     \begin{tabular}{c}
\epsfig{file=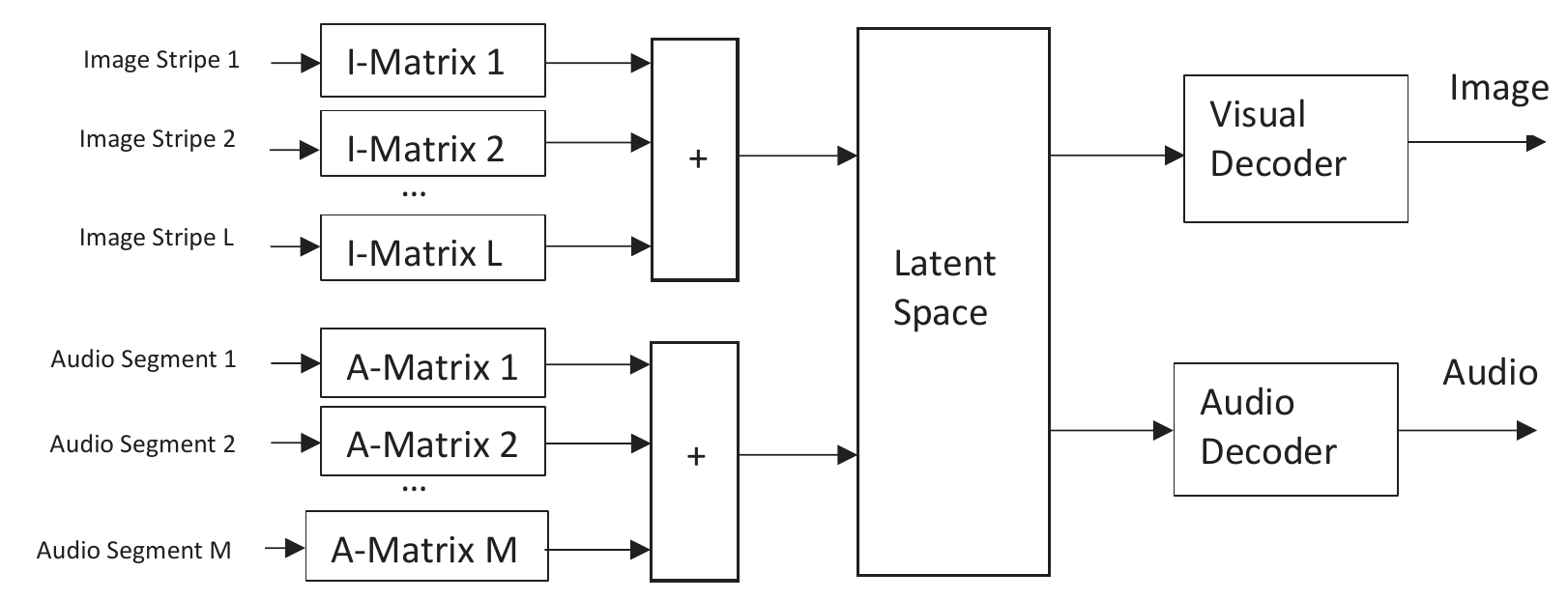,width=5.3in}  
 \end{tabular}
    \end{center}
    \caption{
    The architecture of PC-VAE with multimodal Transformer. }
    \label{fig:arch}
\end{figure}

We further extend the transformer to multimodal transformer, which can split any multimodal data  to a number of smaller-size single modality data (e.g., images or audio segments). 
We propose a parallel concatenated architecture for multimodal transformer in PC-VAE, as shown in Fig. \ref{fig:arch}. It illustrates the PC-VAE with two modalities, image and audio.

The RGB image stripe $\bf x$ is a matrix, then we can
vectorize it \cite{Barr18},
which is the concatenation of its columns, i.e., $vec({\bf x})=[x_1,x_2,\cdots,x_N]^t$. 
For image stripes, each of the parallel concatenated transformer can be expressed as
\begin{equation}
{\bs \mu}_i=\Phi_i {\bf x}_i
\label{equ:comp2}
\end{equation}
where $i=1,2,\cdots, L$, and $\Phi_i$ is compression matrix having size $N_1'\times N_1$ and $N_1'< N_1$.
${\bs \mu}_i$ is the output of compression for image signals, and it has length 
$N_1'$, versus the original length $N_1$ of ${\bf x}_i$.
In Fig. \ref{fig:arch}, $\Phi_i$ is Image-matrx (I-matrix) $i$ for image stripes.
Then all the ${\bs\mu}_i$ ($i=1,2,\cdots,L$) values are summed together, we get
\be
{\bs\mu_v}=\sum_{i=1}^L {\bs \mu}_i
\ee

Regarding matrix for audio segments in Fig. \ref{fig:arch}, we can choose different matrix $\Psi_j$,
and 
\begin{equation}
{\bs \mu}_j=\Psi_j {\bf y}_j
\label{equ:comp3}
\end{equation}
where 
${\bf y}_j$ is audio segment and $j=1,2,\cdots, M$.
If the size of matrix $\Phi_j$ is 
$N_2'\times N_2$ ($N_2'< N_2$), then the output ${\bs \mu}_j$ has length $N_2'$.
Then all the ${\bs\mu}_j$ ($j=1,2,\cdots,M$) values are summed together, we get
\be
{\bs\mu_a}=\sum_{i=1}^M {\bs \mu}_j
\ee

In this paper, we choose the compression matrices $\Phi_i$
and $\Psi_j$ 
as zero-mean Gaussian random matrix with unit variance.
The compression matrix $\Phi_i$ and $\Psi_j$ are generated randomly, but once they are generated, all values in the matrix are frozen during the training and testing process.
This is quite different with the current KL-VAE in which the parameters in the encoder neural network need to be tuned in the training process. Our approach saves computation because it doesn't need any training in the encoders.

Then we can compute the variance $\sigma_v^2$ of all visual signals, 
and a reparameterization process is performed based on ${\bs \mu}_v$ and $\sigma_v$,
\begin{equation}
{\bs z}_v={\bs \mu}_v+\sigma_v {\bs \epsilon}
\label{equ:rep2}
\end{equation}
where ${\bs \epsilon}\sim {\mathcal N}({\bs 0, I})$.
Similarly, we can use 
reparameterization process to get 
${\bs z}_a$ 
based on ${\bs \mu}_a$ and $\sigma_a$
where $\sigma_a^2$ is the variance of all audio signals.
\begin{equation}
{\bs z}_a={\bs \mu}_a+\sigma_a {\bs \epsilon}
\label{equ:rep3}
\end{equation}

The latent space in Fig. \ref{fig:arch} is a vector, and it is obtained based on the serial concatenation of ${\bf z}_v$ and ${\bf z}_a$. So the total length of latent space in Fig. \ref{fig:arch}
is $N_1'+N_2'$.
The latent space $\bs z$ serves as the input to the PC-VAE two decoders, visual decoder and audio decoders.

The visual and audio decoders are just two convolutional neural networks (CNN).
Their inputs are from the latent space $\bs z$, but their outputs are different.
The visual decoder has colored image output, and the audio decoder produces audio signals. For different applications, the CNN structures are different.

The benefits of our multimodal transformer includes the following.
\begin{enumerate}
    \item 
    The vectorization of images are performed in columns, and our vision transformer partitions the images into stripes by columns, which keeps the original spatial relations of pixels.
    
    \item
    The Gaussian random matrices $\Phi_i$ and $\Psi_j$ are chosen randomly, and they don't need any selection and training process.

    \item
    All the operations of latent space construction are linear, so it saves computational cost.
    
    \item
    The inputs to the visual decoder and audio decoder are the same, which makes data processing easier.
    
    \item
    All the linear transformation in the multimodal transformer and the neural network processing in the decoders are in parallel, which can tremendously increase the processing speed.
    
\end{enumerate}

\subsection{Loss Function Using Interaction Information}
\label{sect:2}

\begin{figure}[ht]
    \begin{center}
     \begin{tabular}{c}
\epsfig{file=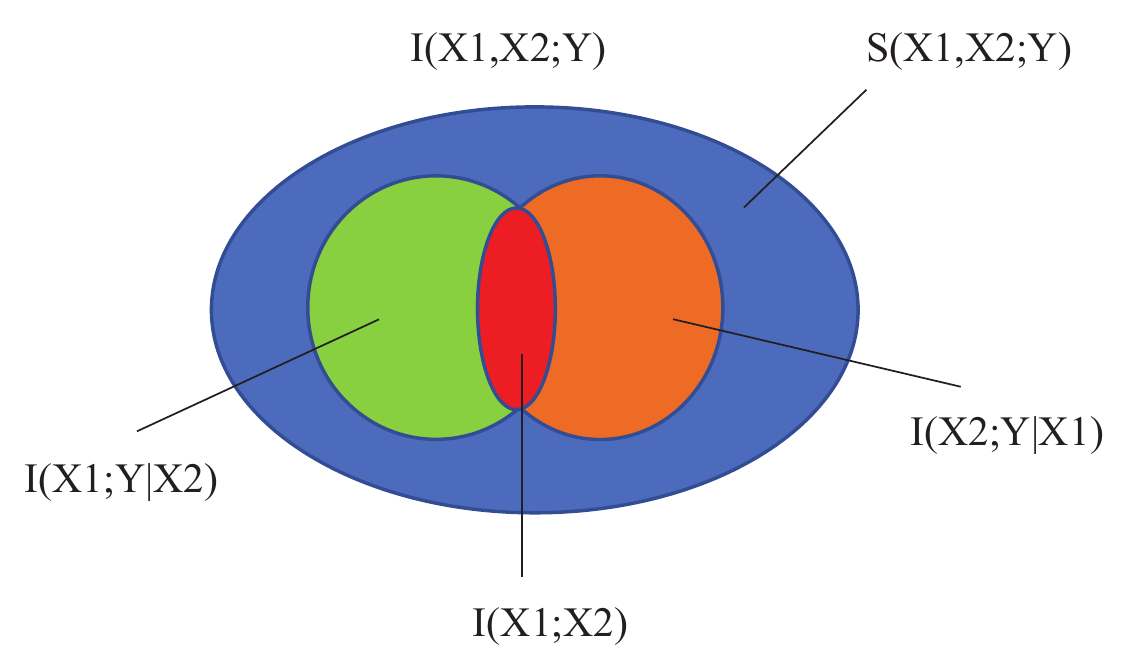,width=2.5in}  
 \end{tabular}
    \end{center}
    \caption{
    An illustration of PID with four parts~\cite{Will10}. }
    \label{fig:1}
\end{figure}

Partial Information Decomposition (PID) was proposed to evaluate 
the mutual information between multiple variables~\cite{Will10}~\cite{Niu19}~\cite{Dutt22}.
PID can decompose mutual information to unique information, redundant information, and synergetic information. 
In our cross-modal information generative modeling,
the mutual information between input cross-modalities $(X_1,X_2)$ and output modality $Y$ can be decomposed as~\cite{Sun15}\cite{Will10}
\bea
I(X_1,X_2;Y)&=&U(X_1;Y|X_2)+U(X_2;Y|X_1)
+R(X_1,X_2;Y)+S(X_1,X_2;Y)
\eea
where $U(X_1;Y|X_2)$ denotes the unique information of $Y$ present only in $X_1$ and not in $X_2$; 
$U(X_2;Y|X_1)$ denotes the unique information of $Y$ present only in $X_2$ and not in $X_1$; 
$R(X_1,X_2;Y)$ denotes the redundant information of $Y$ present in both $X_1$ and $X_2$;  $S(X_1,X_2; Y)$ is the synergetic information of $Y$ that is not present in $X_1$ or $X_2$ individually, but present in $(X_1, X_2)$ jointly. 
In Fig. \ref{fig:1}, we illustrate the relations of these four parts~\cite{Will10}. The whole region is $I(X_1,X_2;Y)$; the red region is redundant information $R(X_1;X_2)$;
the blue region is synergetic information $S(X_1,X_2;Y)$;  the green and orange regions are the unique information  $U(X_1;Y|X_2)$ and $U(X_2;Y|X_1)$.
Mathematically, they can be represented as~\cite{Will10}
\bea
U(X_1;Y|X_2)&=&I(X_1;Y|X_2)\\
U(X_2;Y|X_1)&=&I(X_2;Y|X_1)\\
R(X_1,X_2;Y)&=&I(X_1;X_2)\\
S(X_1,X_2;Y)&=&I(X_1,X_2;Y)-I(X_1;Y|X_2)-I(X_2;Y|X_1)-I(X_1;X_2)
\label{equ:syn}
\eea
where $I(X_1; X_2)$ denotes the mutual information between $X_1$ and $X_2$, and is defined as~\cite{Cove06},
\be
I(X_1;X_2)=\sum_{x_1\in X_1}\sum_{x_2\in X_2} p(x_1,x_2)\log \frac{p(x_1,x_2)}{p(x_1)p(x_2)}
\ee
It is well known that $I(X_1;X_2)$ is always nonnegative.
$I(X_1;Y|X_2)$ is conditional mutual information~\cite{Cove06}
\be
I(X_1;Y|X_2)=\mathbb{E}_{p(x_1,y,x_2)} p(x_1,y|x_2)\log \frac{p(x_1,y|x_2)}{p(x_1|x_2)p(y|x_2)}
\ee
where $\mathbb{E}$ denotes mathematical expectation.

PID reflects the relations between the four parts. Williams and Beer further defined Interaction Information (II)~\cite{Will10}~\cite{Wang21}~\cite{Sun15},
\bea
II(X_1,X_2;Y)&=&I(X_1;Y|X_2)-I(X_1;Y)
\label{equ:0}
\eea
Based on chain rule~\cite{Cove06},
\be
I(X_1,X_2;Y)=I(X_2;Y)+I(X_1;Y|X_2)
\ee
so (\ref{equ:0}) becomes
\bea
II(X_1,X_2;Y)
&=&I(X_1,X_2;Y)-I(X_2;Y)-I(X_1;Y) 
\label{equ:9} \\
&=&I(X_1,X_2;Y)-[I(X_2;Y|X_1)+I(X_1;X_2)]\nonumber\\
&&
-[I(X_1;Y|X_2)+I(X_1;X_2)] 
\label{equ:10} \\
&=&I(X_1,X_2;Y)-I(X_1;Y|X_2)
-I(X_1;Y|X_2)-2I(X_1;X_2)\\
&=&
[I_{\mv}(X_1,X_2; Y)
-I_{\mv}(X_1; Y|X_2)-I_{\mv}(X_2; Y|X_1)
-I_{\mv}(X_1; X_2)]\nonumber\\
&&-I_{\mv}(X_1; X_2)
\label{equ:11}\\
&=&S(X_1,X_2;Y)-I(X_1;X_2)
\label{equ:12}
\eea
where the derivation from (\ref{equ:9}) to (\ref{equ:10})
is based on the relation~\cite{Will10} 
\be
I(X_2;Y)=I(X_2; Y|X_1)+I(X_1; X_2)
\ee
and from (\ref{equ:11}) to (\ref{equ:12})
is based on the synergetic information in (\ref{equ:syn}).

Result in (\ref{equ:12}) shows that Interaction Information $II(X_1, X_2; Y)$ is the difference between Synergetic Information and Redundancy Information.
It could have the following two cases:
\begin{enumerate}
\item
When $II(X_1,X_2;Y)>0$, 
it means the synergetic information is higher than the redundancy information.

 \item
When $II(X_1,X_2;Y)<0$, it means the redundancy information is higher than the synergetic information. 
\end{enumerate}
We incorporate the interaction information into the synthetic data evaluation and cross modal variational autoencoder training.
We try to minimize the interaction information because the smaller interaction information means smaller synergetic information, and the output $Y$ is more close to the input $X_1$ and $X_2$.

We propose a new loss function based on the above analysis.
Assume the image and audio input to PC-VAE are $X_1$, $X_2$, its output is $Y$,
and $X_1$ has the same modality with $Y$ without losing of generality.
\begin{eqnarray}
&&{\mathcal L}({\bs \theta,\bs \phi,\bf x})\nonumber \\
&=&II(X_1,X_2;Y)
+E(||Y-X_1||^2) \label{equ:errors2}\\
&=&
I(X_1,X_2;Y)-I(X_2;Y)-I(X_1;Y) 
+E(||Y-X_1||^2)
\label{equ:loss2}
\end{eqnarray}
It includes two parts, the interaction information and reconstruction loss.
We will train the PC-VAE via minimizing this loss function.

\section{Experiments}

We ran our simulation on PC-VAE for visual-audio generative modeling based on a Subset from University of Rochester Music Performance (URMP) dataset (Sub-URMP)~\cite{Chen171}\cite{Li16}.
The Sub-URMP contains images and audios cut from the URMP. 
A sliding window with time duration 0.5 seconds and a stride 0.1 second was used to obtain the data. 
The first frame of each video chunk, an image with size $1080\times 1920$, was used to represent the visual content of the sliding window.
The audio files are in WAVE format in stereo channel with a sampling rate of 44 KHz and bit depth of 16 bits~\cite{Chen22}.

 The training set has 71230 paired samples where each pair has one image and and one audio file. The validation set has 9575 paired samples
 We chose 50000 paired samples for training, and 8000 paired samples for testing.
 To make it work more efficiently, we performed down-sampling by 34 for the images, and down-sampled by 10 for the audio files. Since the two channels audio data are identical, we only chose one channel data.  
 Based on this data pre-processing, 
 each image has a size of $32\times 32\times 3$ where $3$ stands for the RGB channels.
 The each audio file has a size of $2205$ which is a 1-D vector.

In our experiment, the PC-VAE encoder is designed with $L=6$ for image stripes, and $M=5$ for audio segments. We chose matrix $\Phi_i$ with the number of columns as $\frac{32\times 32\times 3}{L}=512$, and matrix $\Psi_j$, with the number of columns as $\frac{2205}{M} =441$. The number of rows in $\Phi_i$ and $\Psi_j$ 
could be different in simulations.

The configurations of our PC-VAE two decoders are specified as follows.
In the visual decoder, a CNN with 3 ReLu layers and
4 transposed convolutional layers was used.
The input layer has size of the latent space length.
A ReLu layer comes after a 2-D convolutional layer,
and then another 2-D convolutional layer followed the ReLu layer.
The four convolutional layers have filter size of
$7\times 64$, 
$3\times 64$, 
$3\times 32$, and 
$3\times 3$ respectively. The stride sizes 
are 8, 2, 2, and 1 for the above four layers.  
The output size is 
$32\times 32\times 3$, so that it can match the size of the input image.
The audio decoder has similar structure as the visual decoder,
except for the last layer where we used a fully connected layer with size $2205$ instead of a 2D convolutional layer with size $3\times 3$.

The performance of the multimodal transformed-based PC-VAE for visual generation was evaluated based on two experiments. 1)
when both images and audio are present in the encoder side;
2) cross-modal data generation based on one modality only.

\begin{figure}[ht]
    \begin{center}
     \begin{tabular}{cc}
\epsfig{file=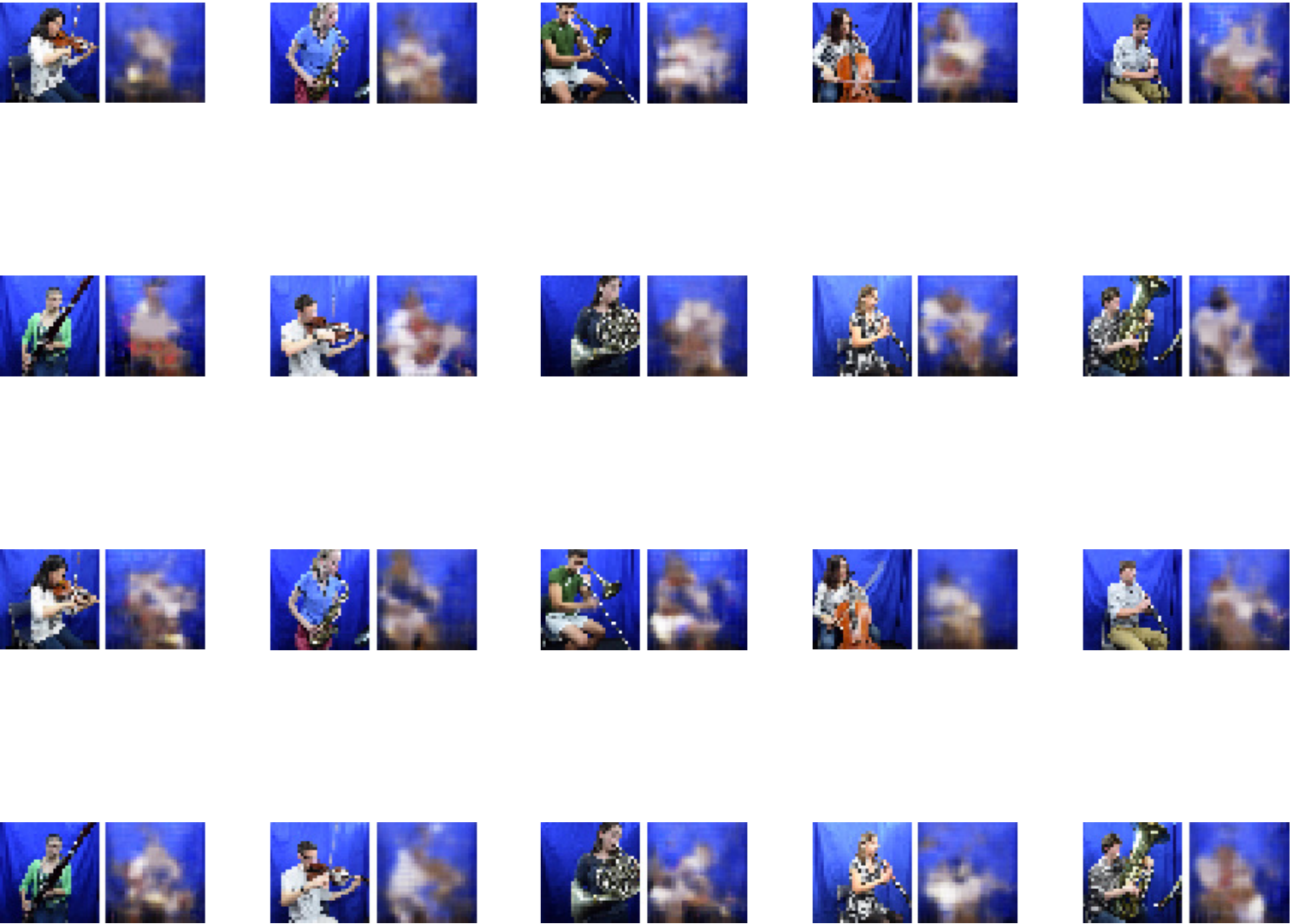,width=2.5in}  &
\epsfig{file=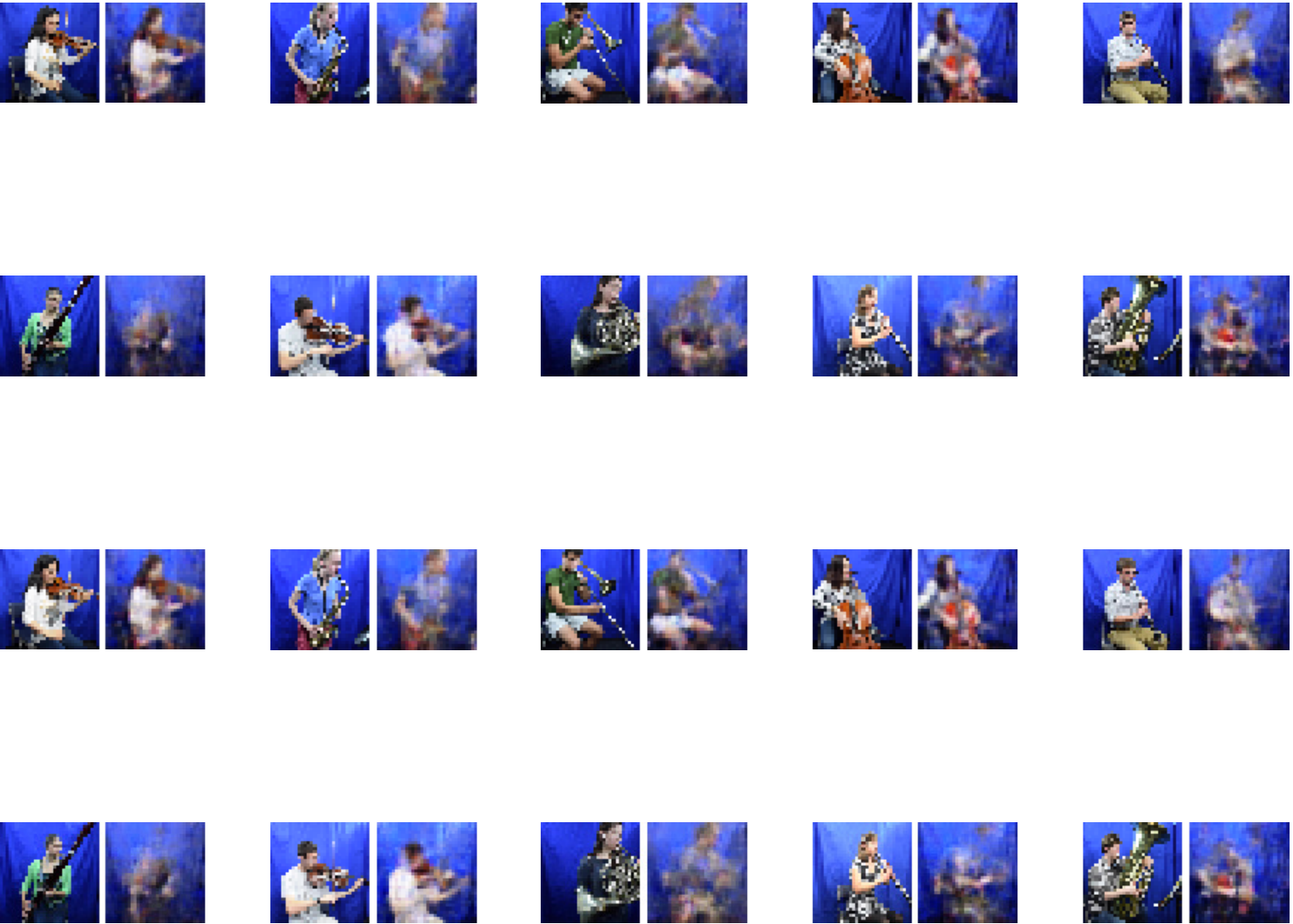,width=2.5in}  \\
(a) & (b)
      \end{tabular}
    \end{center}
    \caption{
Images generated using PC-VAE based on audio information only.
(a) Latent Space length=200; (b) Latent Space length=250.
}
    \label{fig:audio}
\end{figure}

 In the first experiment, both image and audio information was available. The visual and audio encoders outputs were combined into the latent space.
 In Fig. \ref{fig:audio}a, we plot 
 an example image generated by the visual decoder with training for 50 epochs, and the latent space $\bs z$ has length 300 (150 from image stripes and 150 from audio segments). 
 In each pair of images, the left image is the input image, and the right image is an example output from the visual decoder. The output images are very similar to the input image.
 In Fig. \ref{fig:audio}b, we plot 
 an example image generated by the visual decoder with latent space $\bs z$ has length 400 (200 from image stripes and 200 from audio segments).

\begin{figure}[ht]
    \begin{center}
     \begin{tabular}{cc}
\epsfig{file=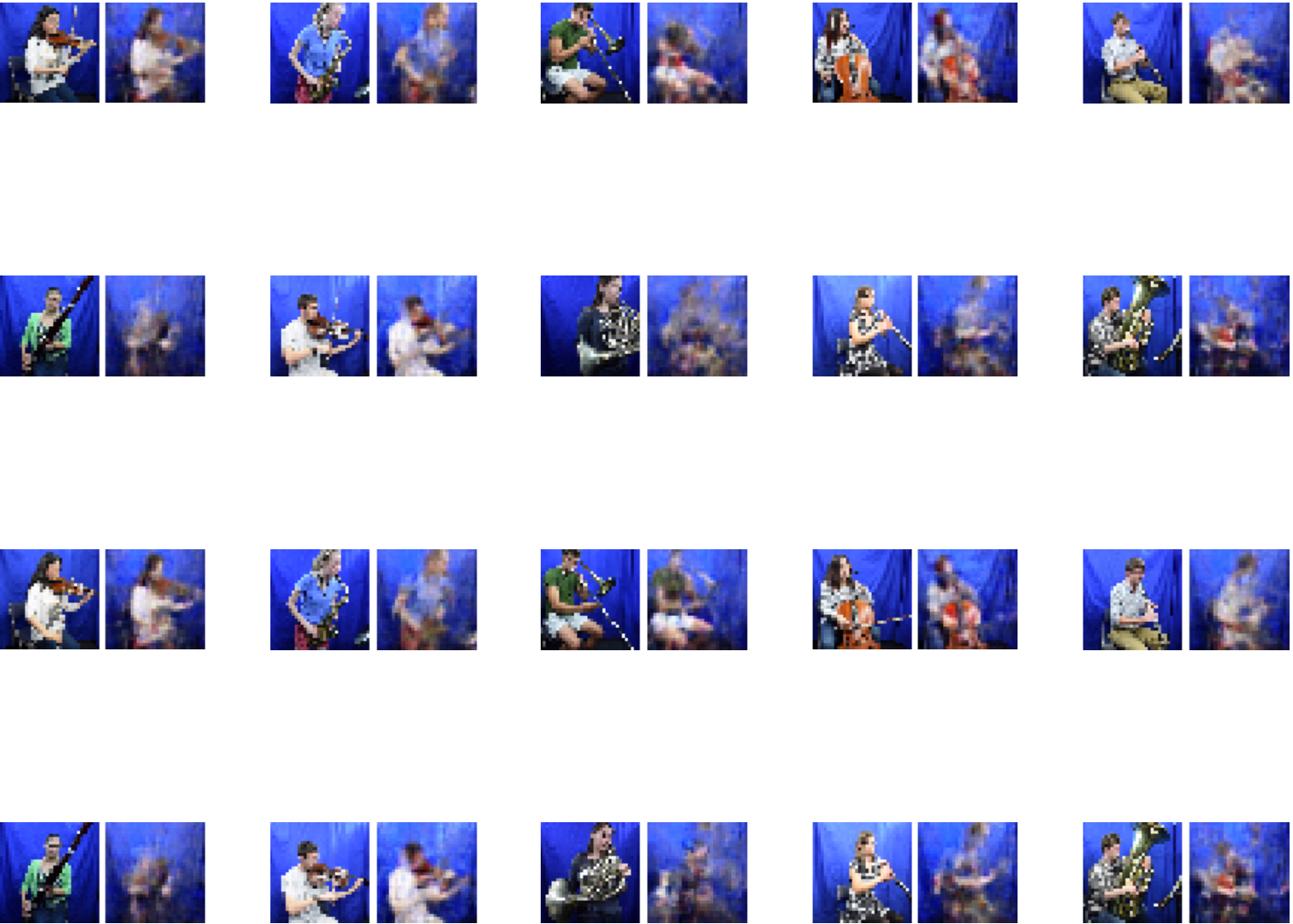,width=2.5in}  &
\epsfig{file=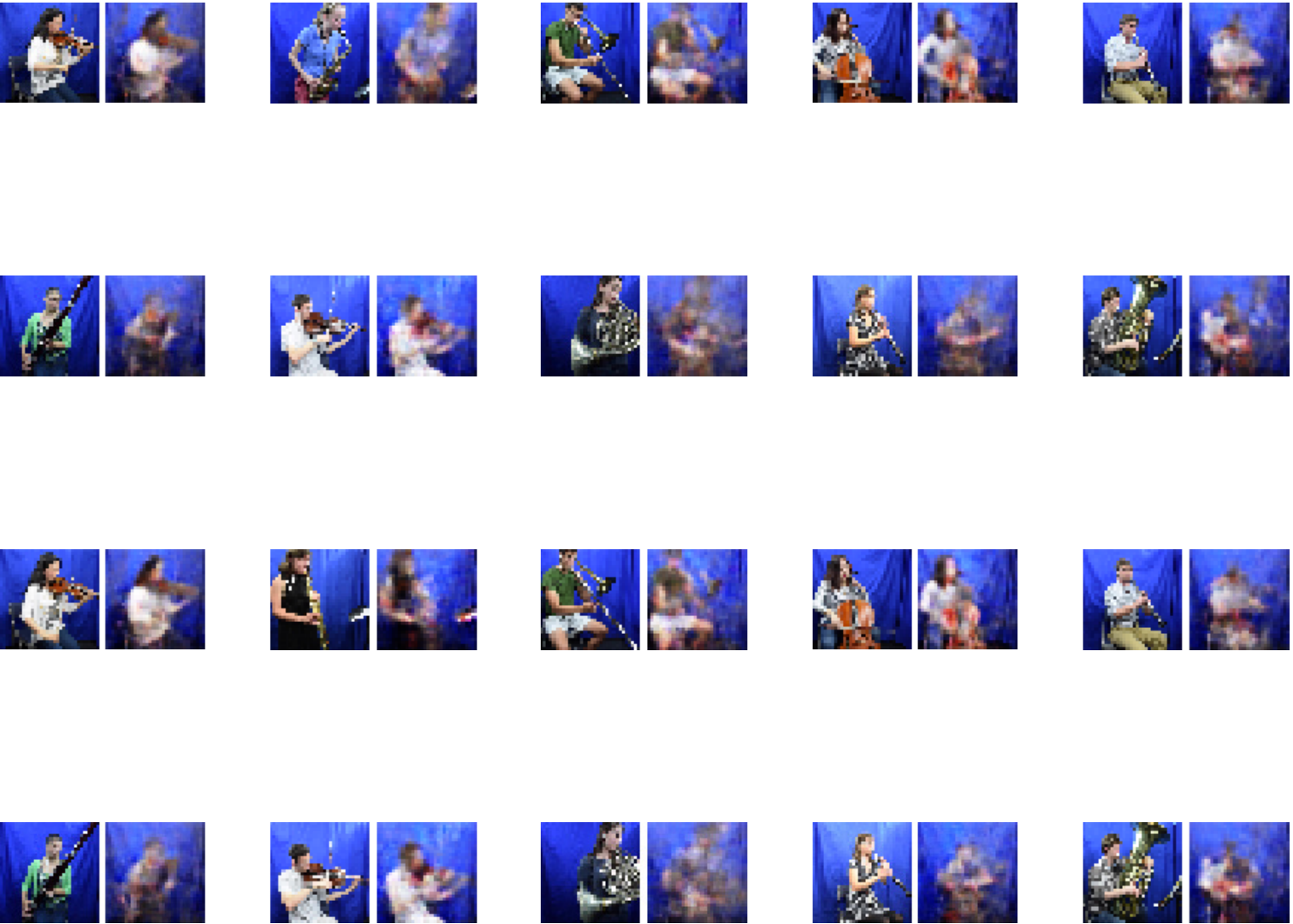,width=2.5in}  \\
(a) & (b)
      \end{tabular}
    \end{center}
    \caption{
Images generated using PC-VAE
based on visual and audio information using loss function in (\ref{equ:loss2}).
(a) Latent Space length=300; (b) Latent Space length=400.
}
    \label{fig:vis_audio}
\end{figure}

\begin{figure}[ht]
    \begin{center}
     \begin{tabular}{c}
\epsfig{file=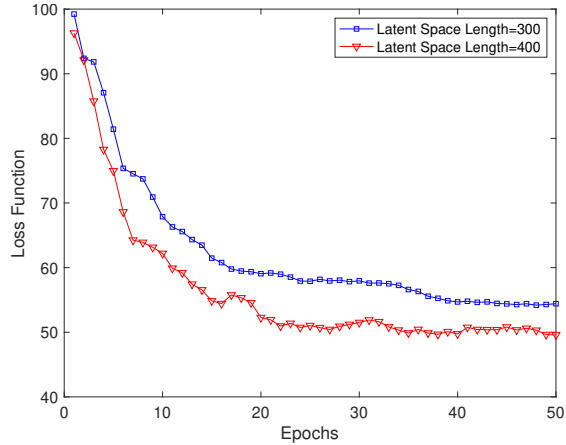,width=2.9in}  
      \end{tabular}
    \end{center}
    \caption{
Reconstruction errors in multimodal transformer-based PC-VAE for visual generation with loss function in (\ref{equ:loss2}).  
}
    \label{Fig:perf}
\end{figure}

In Fig. \ref{Fig:perf}, we plot the loss function in (\ref{equ:loss2}) based on the interaction information 
for the experiment when both visual and audio information were present. Observe that the loss function (\ref{equ:loss2}) reduces almost monotonically versus the training epoches.

In the second experiment, the visual information was missing, and only audio information was available in the validation, and the images were generated using audio input. 
  In Fig. \ref{fig:audio}a, we plot an example image generated by the visual decoder with training for 50 epochs, and the latent space has length 200. In each pair of images, the left image is the missing image (blind to the encoder and decoder), and the right image is an example output from the visual decoder. 
 Observe that visual signal could be generated based on audio signal only.
 Although we only present one example, the visual decoder could generate any number of images based on one audio input.
 In Fig. \ref{fig:audio}b, we plot an example image generated by the visual decoder and the latent space has length 250.

\begin{figure}[ht]
    \begin{center}
     \begin{tabular}{cc}
\epsfig{file=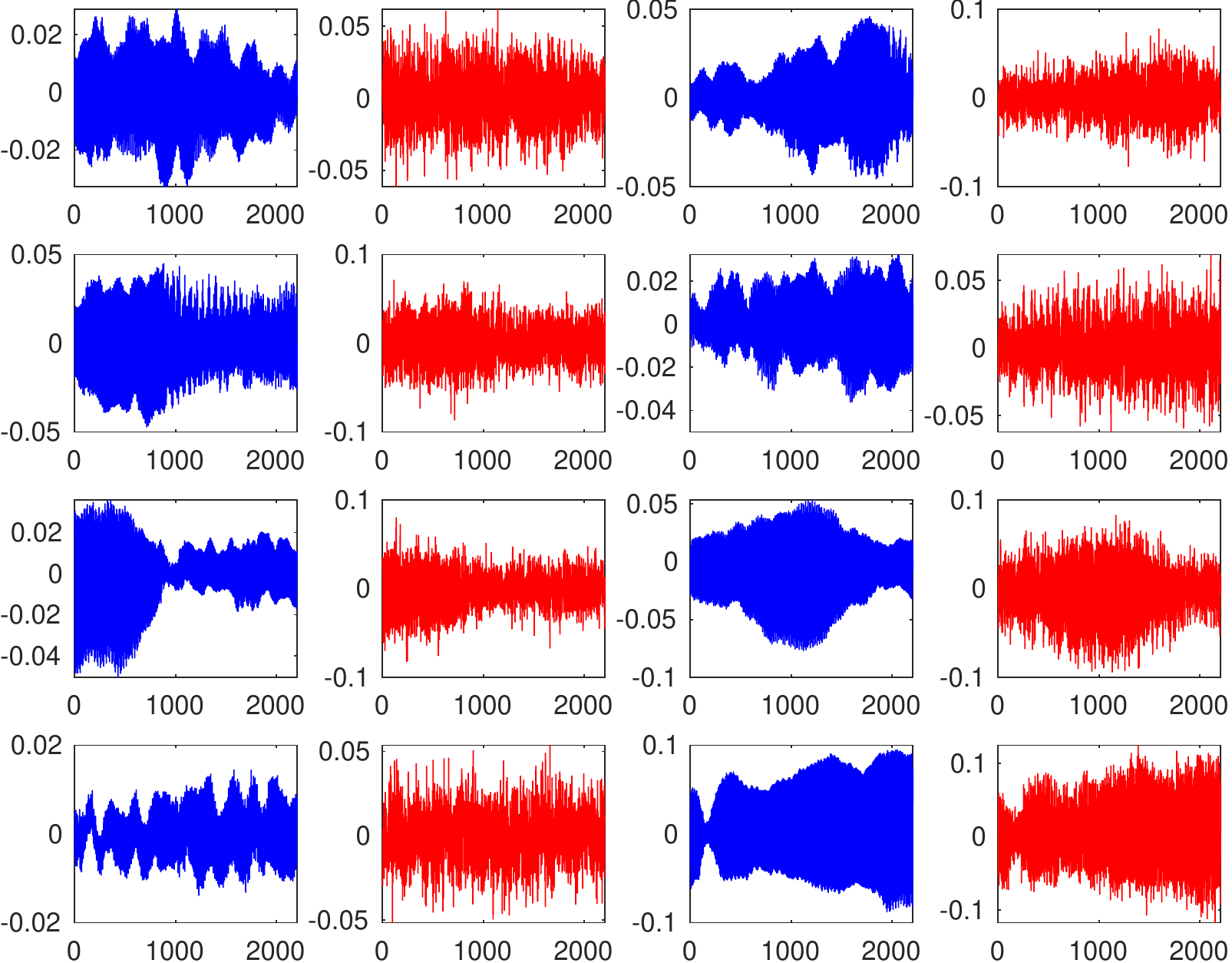,height=2.in}  &
\epsfig{file=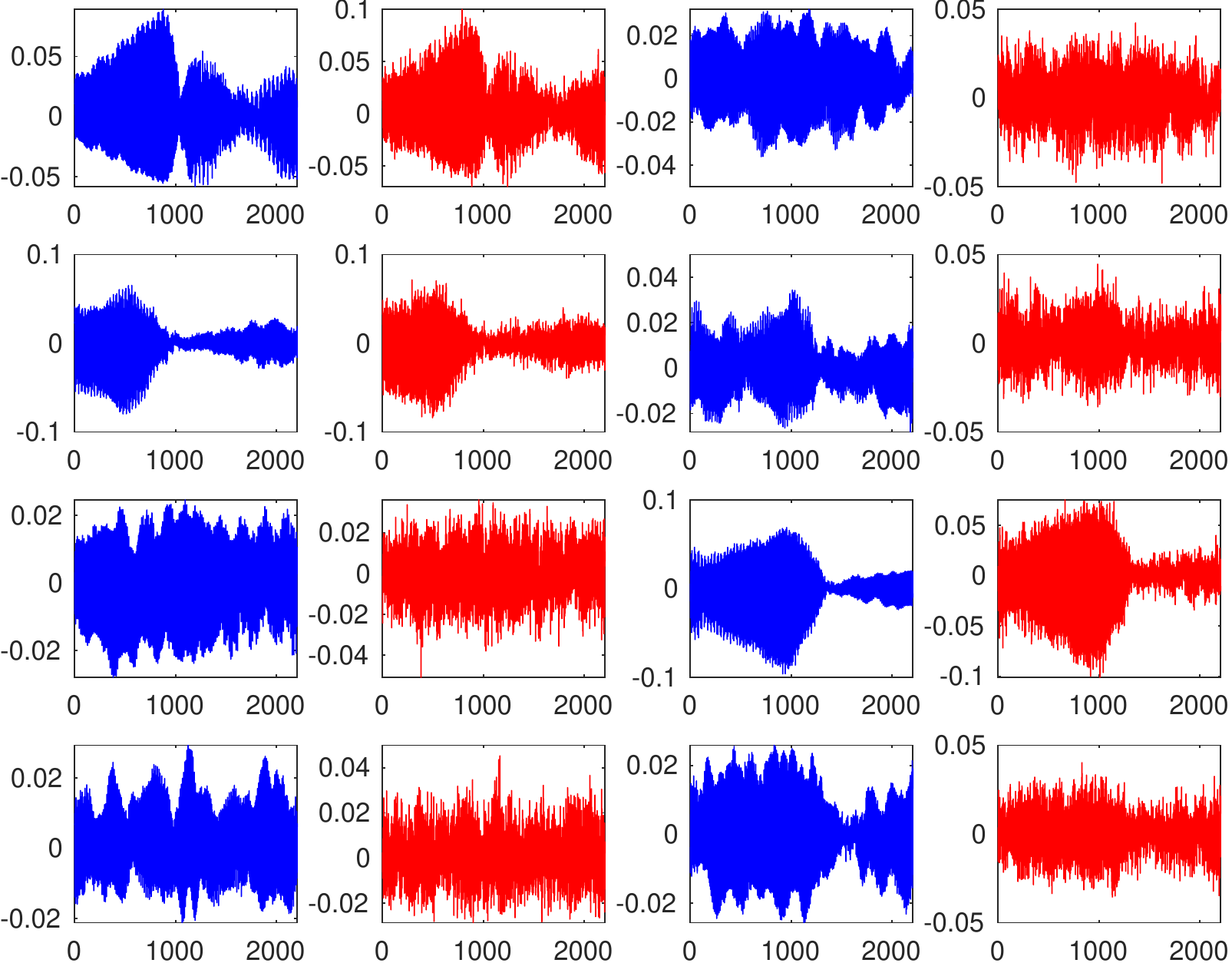,height=2.in}  \\
(a) & (b)
      \end{tabular}
    \end{center}
   \caption{Original audio signals (in blue) versus the generated audio signal (in red). (a) The audio signals were generated using visual only when audio input was missing. 
   (b) Audio signals generated based on visual and audio input.
    }
    \label{Fig:visual4audio}
\end{figure}

 The performance of the PC-VAE for audio generation was evaluated based on two experiments. In the first experiment, the audio information was missing, and only image information was available in the validation, and the audio signals were generated using visual input. 
 In Fig. \ref{Fig:visual4audio}a, we plot an example audio signals generated by the audio decoder with training for 50 epochs, and the latent space has length 200. In each pair of audio signals, the left plot in blue is the missing audio (blind to the encoder and decoder), and the right plot (in red) is an example output from the audio decoder. Of course, the audio decoder could generate any number of audio plot based on one visual input.
 Observe that audio signal could be generated based on visual signal only.
 In the second experiment, both visual and audio information were available. 
 In Fig. \ref{Fig:visual4audio}b, we plot 
 an example audio generated by the audio decoder with training for 50 epochs, and the latent space has length 400 (200 from visual latent space and 200 from audio latent space). In each pair of plot (blue and red), the left plot in blue is the input audio, and the right plot is an example output from the audio decoder. Observe that the output audio plots are very similar to the input audio signals, and the audio signals could be generated successfully in both cases.

\section{Conclusions}

 We have proposed a multimodal transformer using parallel architecture for 
     for synthetic multimodal data generation. 
The multimodal transformer is designed using multiple compression matrices.
We propose a new vision transformer. Instead of using patches, we use column stripes for images in R, G, B channels. The column stripes keep the spatial relations of original image.
We have proposed a new machine learning model, PC-VAE.
The multimodal transformer serves as encoders for PC-VAE.
The PC-VAE consists of multiple encoders, one latent space, and two decoders (visual and audio).
Each matrix in the multimodal transformer is chosen as zero-mean unitary Gaussian random matrix.

We have proposed a new cost function based on the interaction information from partial information decomposition.
The interaction information evaluates the input cross-modal information and 
decoder output. 
The parallel concatenated variational autoencoders are trained 
via minimizing the interaction information.
Experiments are performed to validate the proposed multimodal transformer for parallel concatenated VAEs.

The multimodal transformed-based PC-VAE has potential to help disabled persons to sense the world using the modality that they are capable to perceive. 
For example, visual signals could be transformed to audible signals using the PC-VAE, so that a visually-impaired person can sense what other people see.
A piece of music may not make any sense to hearing-impaired persons, but PC-VAE can transform it to images or video for them to view. 

\newpage

\bibliographystyle{plain} 
\bibliography{NIPS} 

\begin{thebibliography}{10}

\bibitem{akbari2021vatt}
Hassan Akbari, Liangzhe Yuan, Rui Qian, Wei-Hong Chuang, Shih-Fu Chang, Yin
  Cui, and Boqing Gong.
\newblock Vatt: Transformers for multimodal self-supervised learning from raw
  video, audio and text.
\newblock {\em Advances in Neural Information Processing Systems},
  34:24206--24221, 2021.

\bibitem{Balt18}
Tadas Baltru{\v{s}}aitis, Chaitanya Ahuja, and Louis-Philippe Morency.
\newblock Multimodal machine learning: A survey and taxonomy.
\newblock {\em IEEE transactions on pattern analysis and machine intelligence},
  41(2):423--443, 2018.

\bibitem{Barr18}
Shane Barratt.
\newblock A matrix gaussian distribution.
\newblock {\em arXiv preprint arXiv:1804.11010}, 2018.

\bibitem{Chen171}
Lele Chen, Sudhanshu Srivastava, Zhiyao Duan, and Chenliang Xu.
\newblock Deep cross-modal audio-visual generation.
\newblock In {\em Proceedings of the on Thematic Workshops of ACM Multimedia
  2017}, pages 349--357, 2017.

\bibitem{chen2021history}
Shizhe Chen, Pierre-Louis Guhur, Cordelia Schmid, and Ivan Laptev.
\newblock History aware multimodal transformer for vision-and-language
  navigation.
\newblock {\em Advances in Neural Information Processing Systems},
  34:5834--5847, 2021.

\bibitem{cheng2021multimodal}
Junyan Cheng, Iordanis Fostiropoulos, Barry Boehm, and Mohammad Soleymani.
\newblock Multimodal phased transformer for sentiment analysis.
\newblock In {\em Proceedings of the 2021 Conference on Empirical Methods in
  Natural Language Processing}, pages 2447--2458, 2021.

\bibitem{Cove06}
Thomas~M Cover.
\newblock {\em Elements of information theory}.
\newblock John Wiley \& Sons, 1999.

\bibitem{Doso21}
Alexey Dosovitskiy, Lucas Beyer, Alexander Kolesnikov, Dirk Weissenborn,
  Xiaohua Zhai, Thomas Unterthiner, Mostafa Dehghani, Matthias Minderer, Georg
  Heigold, Sylvain Gelly, et~al.
\newblock An image is worth 16x16 words: Transformers for image recognition at
  scale.
\newblock {\em arXiv preprint arXiv:2010.11929}, 2020.

\bibitem{Dutt22}
Sanghamitra Dutta, Praveen Venkatesh, and Pulkit Grover.
\newblock Quantifying feature contributions to overall disparity using
  information theory.
\newblock {\em arXiv preprint arXiv:2206.08454}, 2022.

\bibitem{Dzabraev21}
Maksim Dzabraev, Maksim Kalashnikov, Stepan Komkov, and Aleksandr Petiushko.
\newblock Mdmmt: Multidomain multimodal transformer for video retrieval.
\newblock In {\em Proceedings of the IEEE/CVF Conference on Computer Vision and
  Pattern Recognition}, pages 3354--3363, 2021.

\bibitem{Hao18}
Wangli Hao, Zhaoxiang Zhang, and He~Guan.
\newblock Cmcgan: A uniform framework for cross-modal visual-audio mutual
  generation.
\newblock In {\em Proceedings of the AAAI conference on artificial
  intelligence}, volume~32, 2018.

\bibitem{He21}
Kaiming He, Xinlei Chen, Saining Xie, Yanghao Li, Piotr Doll{\'a}r, and Ross
  Girshick.
\newblock Masked autoencoders are scalable vision learners.
\newblock In {\em Proceedings of the IEEE/CVF Conference on Computer Vision and
  Pattern Recognition}, pages 16000--16009, 2022.

\bibitem{hendricks2021decoupling}
Lisa~Anne Hendricks, John Mellor, Rosalia Schneider, Jean-Baptiste Alayrac, and
  Aida Nematzadeh.
\newblock Decoupling the role of data, attention, and losses in multimodal
  transformers.
\newblock {\em Transactions of the Association for Computational Linguistics},
  9:570--585, 2021.

\bibitem{Hu21}
Ronghang Hu and Amanpreet Singh.
\newblock Unit: Multimodal multitask learning with a unified transformer.
\newblock In {\em Proceedings of the IEEE/CVF International Conference on
  Computer Vision}, pages 1439--1449, 2021.

\bibitem{hu2020iterative}
Ronghang Hu, Amanpreet Singh, Trevor Darrell, and Marcus Rohrbach.
\newblock Iterative answer prediction with pointer-augmented multimodal
  transformers for textvqa.
\newblock In {\em Proceedings of the IEEE/CVF Conference on Computer Vision and
  Pattern Recognition}, pages 9992--10002, 2020.

\bibitem{Huang20}
Jian Huang, Jianhua Tao, Bin Liu, Zheng Lian, and Mingyue Niu.
\newblock Multimodal transformer fusion for continuous emotion recognition.
\newblock In {\em ICASSP 2020-2020 IEEE International Conference on Acoustics,
  Speech and Signal Processing (ICASSP)}, pages 3507--3511. IEEE, 2020.

\bibitem{huang2021unifying}
Yupan Huang, Hongwei Xue, Bei Liu, and Yutong Lu.
\newblock Unifying multimodal transformer for bi-directional image and text
  generation.
\newblock In {\em Proceedings of the 29th ACM International Conference on
  Multimedia}, pages 1138--1147, 2021.

\bibitem{kant2020spatially}
Yash Kant, Dhruv Batra, Peter Anderson, Alexander Schwing, Devi Parikh, Jiasen
  Lu, and Harsh Agrawal.
\newblock Spatially aware multimodal transformers for textvqa.
\newblock In {\em European Conference on Computer Vision}, pages 715--732.
  Springer, 2020.

\bibitem{King14}
Diederik~P Kingma and Max Welling.
\newblock Auto-encoding variational bayes.
\newblock {\em arXiv preprint arXiv:1312.6114}, 2013.

\bibitem{Chen22}
Z.~Duan L.~Chen, S.~Srivastava and C.~Xu.
\newblock Deep cross-modal audio-visual generation.
\newblock {\em https://www.cs.rochester.edu/~cxu22/d/vagan//}, 2022.

\bibitem{le2019multimodal}
Hung Le, Doyen Sahoo, Nancy~F Chen, and Steven~CH Hoi.
\newblock Multimodal transformer networks for end-to-end video-grounded
  dialogue systems.
\newblock {\em arXiv preprint arXiv:1907.01166}, 2019.

\bibitem{lee2020parameter}
Sangho Lee, Youngjae Yu, Gunhee Kim, Thomas Breuel, Jan Kautz, and Yale Song.
\newblock Parameter efficient multimodal transformers for video representation
  learning.
\newblock {\em arXiv preprint arXiv:2012.04124}, 2020.

\bibitem{Li16}
Bochen Li, Xinzhao Liu, Karthik Dinesh, Zhiyao Duan, and Gaurav Sharma.
\newblock Creating a classical musical performance dataset for multimodal music
  analysis: Challenges, insights, and applications.
\newblock {\em IEEE Trans. Multimedia. submitted. Available: https://arxiv.
  org/abs/1612.08727}, 2016.

\bibitem{Li21}
Jing Li, Di~Kang, Wenjie Pei, Xuefei Zhe, Ying Zhang, Zhenyu He, and Linchao
  Bao.
\newblock Audio2gestures: Generating diverse gestures from speech audio with
  conditional variational autoencoders.
\newblock In {\em Proceedings of the IEEE/CVF International Conference on
  Computer Vision}, pages 11293--11302, 2021.

\bibitem{li2021bridging}
Zekang Li, Zongjia Li, Jinchao Zhang, Yang Feng, and Jie Zhou.
\newblock Bridging text and video: A universal multimodal transformer for
  audio-visual scene-aware dialog.
\newblock {\em IEEE/ACM Transactions on Audio, Speech, and Language
  Processing}, 29:2476--2483, 2021.

\bibitem{Lian21}
Stephen~D Liang.
\newblock Variational autoencoder for data analytics in internet of things
  based on transfer entropy.
\newblock {\em IEEE Internet of Things Journal}, 8(20):15267--15275, 2021.

\bibitem{liu2020multi}
Ao~Liu, Shuai Yuan, Chenbin Zhang, Congjian Luo, Yaqing Liao, Kun Bai, and
  Zenglin Xu.
\newblock Multi-level multimodal transformer network for multimodal recipe
  comprehension.
\newblock In {\em Proceedings of the 43rd International ACM SIGIR conference on
  research and development in Information Retrieval}, pages 1781--1784, 2020.

\bibitem{Luca19}
James Lucas, George Tucker, Roger~B Grosse, and Mohammad Norouzi.
\newblock Don't blame the elbo! a linear vae perspective on posterior collapse.
\newblock {\em Advances in Neural Information Processing Systems}, 32, 2019.

\bibitem{Lucas19}
James Lucas, George Tucker, Roger~B Grosse, and Mohammad Norouzi.
\newblock Don't blame the elbo! a linear vae perspective on posterior collapse.
\newblock {\em Advances in Neural Information Processing Systems}, 32, 2019.

\bibitem{ma2022multimodal}
Mengmeng Ma, Jian Ren, Long Zhao, Davide Testuggine, and Xi~Peng.
\newblock Are multimodal transformers robust to missing modality?
\newblock In {\em Proceedings of the IEEE/CVF Conference on Computer Vision and
  Pattern Recognition}, pages 18177--18186, 2022.

\bibitem{Niu19}
Xueyan Niu and Christopher~J Quinn.
\newblock A measure of synergy, redundancy, and unique information using
  information geometry.
\newblock In {\em 2019 IEEE International Symposium on Information Theory
  (ISIT)}, pages 3127--3131. IEEE, 2019.

\bibitem{parthasarathy2021detecting}
Srinivas Parthasarathy and Shiva Sundaram.
\newblock Detecting expressions with multimodal transformers.
\newblock In {\em 2021 IEEE Spoken Language Technology Workshop (SLT)}, pages
  636--643. IEEE, 2021.

\bibitem{rahman2020integrating}
Wasifur Rahman, Md~Kamrul Hasan, Sangwu Lee, Amir Zadeh, Chengfeng Mao,
  Louis-Philippe Morency, and Ehsan Hoque.
\newblock Integrating multimodal information in large pretrained transformers.
\newblock In {\em Proceedings of the conference. Association for Computational
  Linguistics. Meeting}, volume 2020, page 2359. NIH Public Access, 2020.

\bibitem{Sun15}
Jie Sun, Dane Taylor, and Erik~M Bollt.
\newblock Causal network inference by optimal causation entropy.
\newblock {\em SIAM Journal on Applied Dynamical Systems}, 14(1):73--106, 2015.

\bibitem{Tsai19}
Yao-Hung~Hubert Tsai, Shaojie Bai, Paul~Pu Liang, J~Zico Kolter, Louis-Philippe
  Morency, and Ruslan Salakhutdinov.
\newblock Multimodal transformer for unaligned multimodal language sequences.
\newblock In {\em Proceedings of the conference. Association for Computational
  Linguistics. Meeting}, volume 2019, page 6558. NIH Public Access, 2019.

\bibitem{Wang21}
Li-Min Wang, Peng Chen, Musa Mammadov, Yang Liu, and Si-Yuan Wu.
\newblock Alleviating the independence assumptions of averaged one-dependence
  estimators by model weighting.
\newblock {\em Intelligent Data Analysis}, 25(6):1431--1451, 2021.

\bibitem{wang2020transmodality}
Zilong Wang, Zhaohong Wan, and Xiaojun Wan.
\newblock Transmodality: An end2end fusion method with transformer for
  multimodal sentiment analysis.
\newblock In {\em Proceedings of The Web Conference 2020}, pages 2514--2520,
  2020.

\bibitem{Will10}
Paul~L Williams and Randall~D Beer.
\newblock Nonnegative decomposition of multivariate information.
\newblock {\em arXiv preprint arXiv:1004.2515}, 2010.

\bibitem{xie2021robust}
Baijun Xie, Mariia Sidulova, and Chung~Hyuk Park.
\newblock Robust multimodal emotion recognition from conversation with
  transformer-based crossmodality fusion.
\newblock {\em Sensors}, 21(14):4913, 2021.

\bibitem{xie2022mnsrnet}
Wuyuan Xie, Tengcong Huang, and Miaohui Wang.
\newblock Mnsrnet: Multimodal transformer network for 3d surface
  super-resolution.
\newblock In {\em Proceedings of the IEEE/CVF Conference on Computer Vision and
  Pattern Recognition}, pages 12703--12712, 2022.

\bibitem{Yao20}
Shaowei Yao and Xiaojun Wan.
\newblock Multimodal transformer for multimodal machine translation.
\newblock In {\em Proceedings of the 58th annual meeting of the association for
  computational linguistics}, pages 4346--4350, 2020.

\bibitem{yu2020improving}
Jianfei Yu, Jing Jiang, Li~Yang, and Rui Xia.
\newblock Improving multimodal named entity recognition via entity span
  detection with unified multimodal transformer.
\newblock Association for Computational Linguistics, 2020.

\bibitem{Yu19}
Jun Yu, Jing Li, Zhou Yu, and Qingming Huang.
\newblock Multimodal transformer with multi-view visual representation for
  image captioning.
\newblock {\em IEEE transactions on circuits and systems for video technology},
  30(12):4467--4480, 2019.

\bibitem{Zadeh19}
Amir Zadeh, Chengfeng Mao, Kelly Shi, Yiwei Zhang, Paul~Pu Liang, Soujanya
  Poria, and Louis-Philippe Morency.
\newblock Factorized multimodal transformer for multimodal sequential learning.
\newblock {\em arXiv preprint arXiv:1911.09826}, 2019.

\bibitem{zhang2022transformer}
Wei Zhang, Feng Qiu, Suzhen Wang, Hao Zeng, Zhimeng Zhang, Rudong An, Bowen Ma,
  and Yu~Ding.
\newblock Transformer-based multimodal information fusion for facial expression
  analysis.
\newblock In {\em Proceedings of the IEEE/CVF Conference on Computer Vision and
  Pattern Recognition}, pages 2428--2437, 2022.

\bibitem{zhao2022hierarchical}
Bin Zhao, Maoguo Gong, and Xuelong Li.
\newblock Hierarchical multimodal transformer to summarize videos.
\newblock {\em Neurocomputing}, 468:360--369, 2022.

\bibitem{zhu2020enhance}
Ron Zhu.
\newblock Enhance multimodal transformer with external label and in-domain
  pretrain: Hateful meme challenge winning solution.
\newblock {\em arXiv preprint arXiv:2012.08290}, 2020.

\end{thebibliography}

  \end{document}